\documentclass[10pt,twocolumn,letterpaper]{article}

\usepackage[final]{iccv}      
\usepackage{booktabs}
\usepackage{multirow}

\usepackage{arabtex}
\usepackage{utf8}
\usepackage{xcolor}

\setcode{utf8}

\definecolor{iccvblue}{rgb}{0.21,0.49,0.74}
\usepackage[pagebackref,breaklinks,colorlinks,allcolors=iccvblue]{hyperref}

\def\paperID{*****} 
\def\confName{ICCV}
\def\confYear{2025}

\title{AutoSign: Direct Pose-to-Text Translation for Continuous Sign Language Recognition}

\author{Samuel Ebimobowei Johnny\\
Carnegie Mellon University Africa\\
Kigali, Rwanda\\
{\tt\small sjohnny@andrew.cmu.edu}
\and
Blessed Guda \\
Carnegie Mellon University Africa\\
Kigali, Rwanda\\
{\tt\small gblessed@andrew.cmu.edu}
\and
Andrew Blayama Stephen\\
Carnegie Mellon University Africa\\
Kigali, Rwanda\\
{\tt\small abstephe@andrew.cmu.edu}
\and
Assane Gueye \\
Carnegie Mellon University Africa\\
Kigali, Rwanda\\
{\tt\small assaneg@andrew.cmu.edu}
}

\begin{document}
\maketitle
\begin{abstract}Continuously recognizing sign gestures and converting them to glosses plays a key role in bridging the gap between the hearing and hearing-impaired communities. This involves recognizing and interpreting the hands, face, and body gestures of the signer, which pose a challenge as it involves a combination of all these features. Continuous Sign Language Recognition (CSLR) methods rely on multi-stage pipelines that first extract visual features, then align variable-length sequences with target glosses using CTC or HMM-based approaches. However, these alignment-based methods suffer from error propagation across stages, overfitting, and struggle with vocabulary scalability due to the intermediate gloss representation bottleneck. To address these limitations, we propose AutoSign, an \textit{autoregressive decoder-only transformer} that directly translates pose sequences to natural language text, bypassing traditional alignment mechanisms entirely. The use of this decoder-only approach allows the model to directly map between the features and the glosses without the need for CTC loss while also directly learning the textual dependencies in the glosses. Our approach incorporates a temporal compression module using 1D CNNs to efficiently process pose sequences, followed by AraGPT2, a pre-trained Arabic decoder, to generate text (glosses). Through comprehensive ablation studies, we demonstrate that hand and body gestures provide the most discriminative features for signer-independent CSLR. By eliminating the multi-stage pipeline, AutoSign achieves substantial improvements on the Isharah-1000 dataset, achieving an improvement of up to 6.1\% in WER score compared to the best existing method.\end{abstract}

\section{Introduction}
\label{sec:intro}
\begin{figure}
    \centering
    \includegraphics[width=0.9\linewidth]{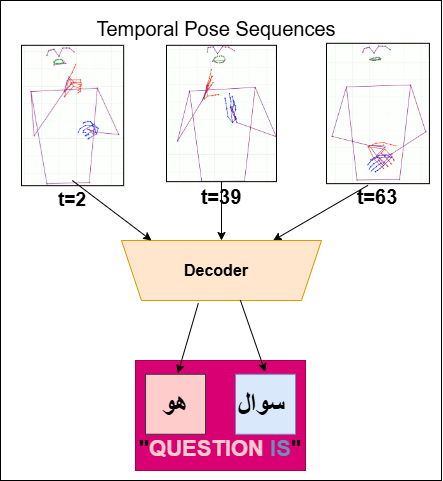}
    \caption{Overview of the proposed DTrOCR method. With the help of the transformer decoder, temporal pose sequences are directly translated to Arabic text without intermediate gloss supervision. This makes the model focus more on end-to-end optimization and cross-modal alignment. To explore the temporal dynamics effectively, a configurable 1D CNN compression is proposed in the pose encoding module.}
    \label{fig:fig1}
\end{figure}

The deaf and hard-of-hearing community, comprising about 5.5\% of the global population \cite{WHODeafness2024}, relies on Sign Language (SL) as their primary mode of communication. Unlike spoken languages, sign languages convey information through a rich combination of hand gestures, facial expressions, and body movements, creating complex visual-spatial languages with separate grammar and vocabulary structures. This intrinsic complexity creates significant challenges for people who are unfamiliar with sign language, limiting effective communication and social inclusion.

While \textit{glosses} remain the fundamental unit of sign language representation, the difference in sign languages across regions and groups presents a significant challenge for widespread adoption of communication technologies. With over 200 distinct sign languages across the world, and each possessing unique gesture signs, it becomes practically impossible for individuals to master multiple SLs. Furthermore, it is unlikely that people without such impairments will learn an additional language, which is not seen as a necessity for them \cite{WFDeafSignLanguageLegalRecognition2020}, further emphasizing the critical need for automated translation systems.

Continuous Sign Language Recognition systems have been primarily developed for popularly used sign languages such as American Sign Language (ASL), Chinese Sign Language (CSL), and German Sign Language (GSL)\cite{luqman2019automatic}.In this work, we focus on developing a robust CSLR system for Saudi Sign Language (SSL) and assessing its performance on unseen signers not included during training. Standard CLSR methods use a multi-stage three-module approach: spatial feature extraction, temporal modeling, and sequence alignment components. These methods typically use Connectionist Temporal Classification (CTC) loss or Hidden Markov Models (HMMs) to map different sequence lengths to the corresponding gloss length, which increases complexity, introduces overfitting, and usually works well for smaller vocabularies\cite{ALYAMI2024103774}.

To address these limitations, we present a novel method that directly translates pose sequences to natural language text without the need for intermediate gloss supervision. In contrast to traditional multi-stage pipelines, our method employs a decoder-only transformer architecture that performs end-to-end autoregressive generation from pose inputs to Arabic text outputs, as illustrated in Figure~\ref{fig:fig1}. This approach eliminates the need for complex alignment mechanisms and gloss annotations while leveraging the robust sequential modeling capabilities of transformer architectures.

Inspired by the success of decoder-only models in language modeling \cite{radford2019language, touvron2023llama}, we demonstrate that autoregressive generation naturally extends to the cross-modal task of pose-to-text translation. Our approach incorporates a configurable temporal compression module using 1D CNNs to efficiently process long pose sequences while preserving important temporal dynamics. Through comprehensive experiments on the Isharah Arabic Sign Language dataset, we demonstrate that direct pose-to-text learning can achieve competitive performance while simplifying the typical CSLR pipeline. Furthermore, we conduct detailed ablation studies analyzing the contribution of different body parts (hands, face, body) to recognition performance, providing insights into optimal pose representations for sign language understanding.

\section{Related Work}
\label{sec:rel}
\subsection{Continuous Sign Language Recognition} Sign language recognition can be categorized into two group: isolated sign language recognition (ISLR) \cite{hu2021hand,9709967,s23042284,10096714,9523142} and continuous sign language recognition (CSLR) \cite{Min_2021_ICCV,9710140,hu2022temporal,hu2023self,hu2023continuous,ALYAMI2025129015,ahn2024slowfast}. Unlike ISLR, which focuses on recognizing individual signs with clear temporal boundaries, CSLR addresses the more challenging task of recognizing continuous sign sequences without prior segmentation, making it essential for real-world applications. Traditional CSLR architectures usually have three main parts: spatial feature extraction, temporal modeling, and sequence alignment \cite{ALYAMI2024103774}. The spatial module extracts frame-level features using 2D-CNNs \cite{10.5555/3600270.3601510,10.1007/978-3-030-58621-8_3,zhou2020spatial}, 3D-CNNs \cite{ahn2024slowfast,10203106}, graph convolutional networks (GCNs) \cite{zhou2020spatial,9534003}, or transformers \cite{Albanie2021bobsl,10378340}. The temporal module captures sequential dependencies through 1D-CNNs \cite{10203106,hu2023continuous,hu2023self,10378475}, spatio-temporal GCNs \cite{9746971,HEDEGAARD2023109528}, RNNs \cite{10203106,10378475,HU2024109903}, or transformers \cite{10.1145/3640815,ALYAMI2025129015}. Finally, the alignment module maps variable-length video sequences to gloss sequences using connectionist temporal classification (CTC) \cite{alyami2025isharahlargescalemultiscenedataset,HU2024109903,cui2023spatial,ahn2024slowfast,zhou2020spatial}, correlation maps \cite{10205442,ALYAMI2024103774}, reinforcement learning \cite{aloysius2020understanding}, or hidden Markov models (HMMs) \cite{8691602,8099847}.
Recent studies have found major problems with these alignment methods: CTC-based approaches often overfit the data and suffer from error propagation across stages, while HMM-based methods typically need to be trained on small vocabularies, which makes them less flexible and harder to apply broadly \cite{ALYAMI2024103774}. 

This three-stage module has become the dominant paradigm in CSLR research, with the most recent focus on improving the alignment components.

\subsection{Pose-based Sign Language Recognition}
 Most CSLR methods \cite{Min_2021_ICCV,9710140,hu2022temporal,hu2023self,ALYAMI2025129015,hu2023continuous,ahn2024slowfast} primarily used RGB video data, but recent approaches have transitioned to pose-based representations\cite{tunga2021pose,s23052853,10378475,9354538,ALYAMI2024103774} derived from frameworks like MediaPipe\cite{lugaresi2019mediapipeframeworkbuildingperception} and OpenPose\cite{8765346}.  The change is driven by several advantages: pose data offers robustness against background variations, lighting circumstances, and clothing discrepancies while preserving critical spatial-temporal information regarding sign articulation.  Additionally, posture representations substantially reduce computational overhead compared to raw RGB processing.

Pose-based models typically extract 2D or 3D keypoints representing body joints, hand landmarks, and facial features, which apply various architectures for temporal modeling. Graph convolutional networks (GCNs) have demonstrated notable accuracy for pose-based CSLR \cite{10378475,9354538,ALYAMI2024103774}, while transformer-based techniques \cite{10.1145/3640815} show promising results. These methods effectively model structural relationships between body joints while capturing temporal changes across frames. Popular CSLR evaluation datasets include RWTH-PHOENIX-Weather 2014 \cite{koller2015continuous}, CSL-Daily \cite{zhou2021improving}, and WLASL \cite{li2020word}. The Isharah dataset \cite{alyami2025isharahlargescalemultiscenedataset} addresses limitations in existing benchmarks by providing uncontrolled environments with various signing conditions.

\subsection{Decoder-Only Architectures}

The success of decoder-only transformers, exemplified by the GPT family, has significantly advanced natural language processing through their superior autoregressive modeling capabilities \cite{radford2019language}. Unlike traditional encoder-decoder architectures that require separate encoding and decoding stages, decoder-only models employ a unified architecture that directly generates target sequences from input contexts in an autoregressive manner. 

Recent work has explored decoder-only architectures for sign language production (text-to-pose) \cite{yin2024t2sgptdynamicvectorquantization}. This method demonstrates the potential of autoregressive generation for sign language tasks, outperforming conventional encoder-decoder and non-autoregressive approaches for text-to-sign production.

However, decoder-only architectures remain unexplored for CSLR (pose-to-gloss). The autoregressive nature of decoder-only models, which has proven highly effective for sequential text generation, presents a natural fit for the temporal sequence modeling required in CSLR. This motivates our investigation of decoder-only transformers for pose-based continuous sign language recognition, marking, to the best of our knowledge, the first work on decoder-only architectures for pose-based CSLR.

\section{Method}

\subsection{Problem Definition}
Let the Isharah dataset consist of $N$ paired examples  
\[
\mathcal{D} = \bigl\{\bigl(\mathbf{X}^{(i)}, \mathbf{Y}^{(i)}\bigr)\bigr\}_{i=1}^N,
\]
where each input $\mathbf{X}^{(i)}\in\mathbb{R}^{T_i\times J\times 2}$ is a frame-wise 2D pose sequence (with $T_i$ frames and $J$ body joints), and each target $\mathbf{Y}^{(i)} = (y^{(i)}_1, \dots, y^{(i)}_{T_Y^{(i)}})$ is the corresponding sequence of Arabic gloss tokens of length $T_Y^{(i)}$.  We group keypoints into four semantic parts: left hand, right hand, face, and body, to enable part-specific data augmentations and embedding strategies.  

Our goal is to learn decoder‑only parameters $\theta$ that model the joint conditional distribution  
\begin{equation}
P\bigl(\mathbf{Y}^{(i)} \mid \mathbf{X}^{(i)}; \theta\bigr)
= \prod_{t=1}^{T_Y^{(i)}}P\bigl(y_t^{(i)} \mid y_{<t}^{(i)},\,\mathbf{X}^{(i)};\theta\bigr)\,,
\end{equation}
where each gloss token is generated autoregressively, conditioned on all previous tokens and the full pose sequence.

\subsection{Overall Architecture}
\begin{figure*}
    \centering
    \includegraphics[width=0.8\linewidth]{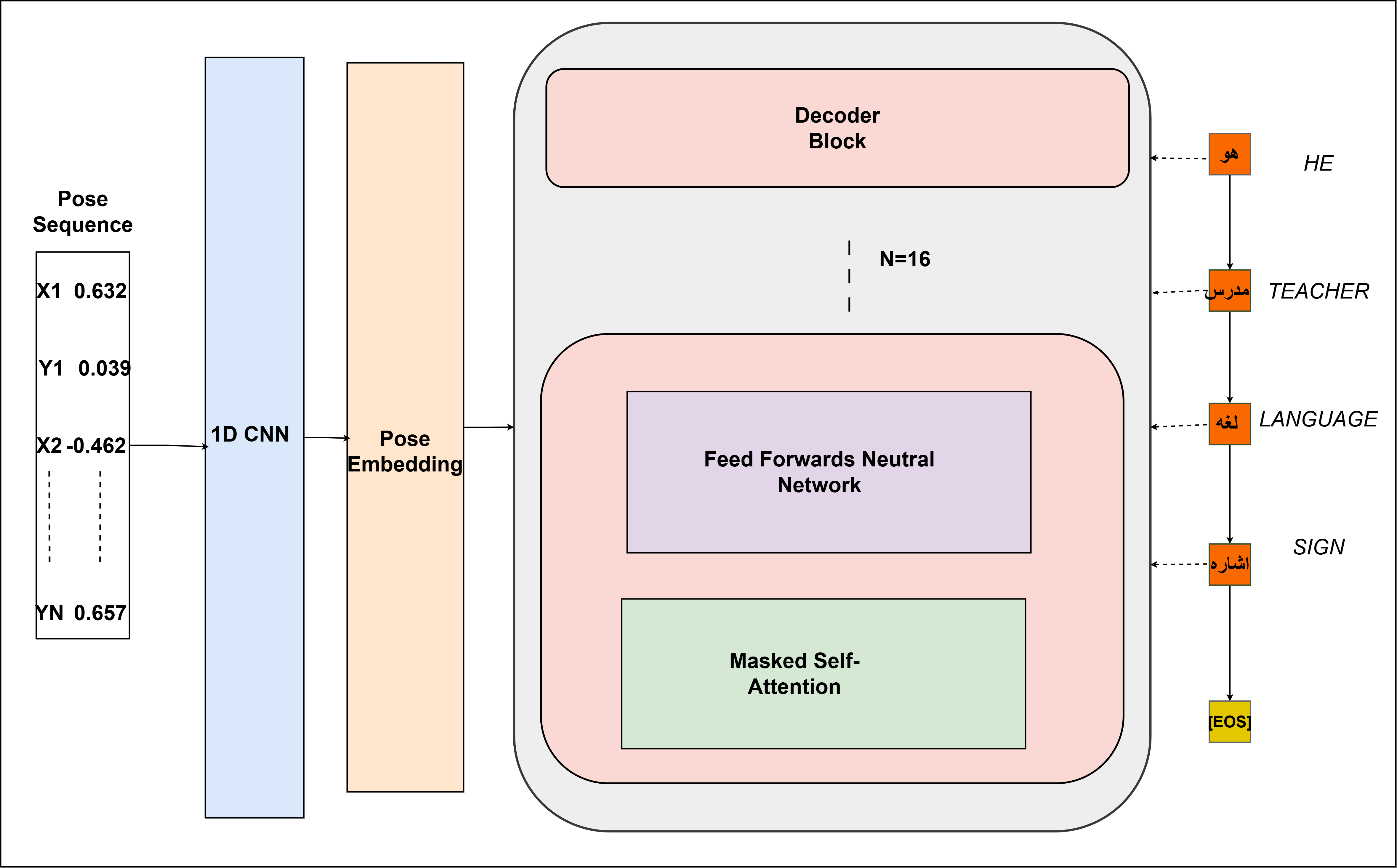}
    \caption{Our proposed Architecture. SL poses are processed through 1D CNN and embedding layers; the encoded representation is fed to a GPT-2 decoder for autoregressive Arabic gloss generation. The model uses pre-trained AraGPT2 initialization and is trained end-to-end on pose-to-text pairs.}
    \label{fig:archi}
\end{figure*}

Figure~\ref{fig:archi} illustrates our \emph{AutoSign} architecture, which adapts the DTrOCR framework \cite{10483819} to sign‑language gloss generation:
\begin{itemize}
  \item \textbf{Pose Encoder (Compression)}  
    A 1D‑CNN module with two convolutional layers (kernel size $3$, stride $2$) compresses the raw pose sequence from $\sim$1000 to $\sim$250 time steps, while expanding the per‑frame feature dimension from $2J{=}134$ to $512$.  
  \item \textbf{Positional Encoding}  
    We inject learnable temporal positional embeddings to each compressed frame embedding to encode sequence order.  
  \item \textbf{Decoder Backbone}  
    We employ AraGPT2 \cite{antoun2020arabert} as a decoder‑only transformer.  Pose embeddings are linearly projected to the GPT‑2 embedding dimension ($768$) and concatenated with gloss token embeddings at each step, then processed through causal self‑attention and cross‑attention layers.  
  \item \textbf{Output Layer}  
    A linear projection followed by softmax over the gloss vocabulary produces the next token probabilities.
\end{itemize}

\subsection{Pose Embedding and Augmentation}
Before compression, we apply part‑aware augmentations to $\mathbf{X}$:
\begin{enumerate}
  \item \emph{Hands}: random rotations and scaling of left/right hand keypoints independently.
  \item \emph{Face}: slight affine jitter to simulate head movement.
  \item \emph{Body}: global pose jitter to model subtle torso shifts.
\end{enumerate}
Augmented keypoints are normalized per sequence and then fed into the CNN compressor.

\subsection{Dynamic Sequence Handling}
Sign sequences vary in length.  At each batch, let $T_{\max} = \max_i T_i$.  All compressed pose sequences are padded with zeros to length $T_{\max}$, and a corresponding binary mask prevents attention to padded frames.  Likewise, gloss sequences are padded on the right to the maximum gloss length in the batch.

\subsection{Training Objective}
We train end‑to‑end with teacher forcing.  The loss for a batch is the sum of cross‑entropy terms only over real (non‑padded) gloss tokens:
\begin{equation}
\mathcal{L}(\theta)
= -\sum_{i=1}^B\sum_{t=1}^{T_Y^{(i)}} 
\log P\bigl(y_t^{(i)} \mid y_{<t}^{(i)},\,\mathbf{X}^{(i)};\theta\bigr)\,,
\end{equation}
where $B$ is the batch size.  We optimize using AdamW with a learning rate of $3\mathrm{e}{-5}$, linear warmup over the first 10\% of steps, and weight‑decay of $0.01$.

\subsection{Implementation Details}
\begin{itemize}
  \item \textbf{Initialization:}  AraGPT2 weights are loaded from a pre-trained checkpoint; pose‑specific layers are randomly initialized.  
  \item \textbf{Batch Size \& Epochs:}  We use $B=32$ and train for 50 epochs, evaluating on a held‑out validation split.  
  \item \textbf{Regularization:}  Dropout ($p=0.1$) is applied in the transformer layers and in the pose compressor.  
\end{itemize}

\section{Experiments}
\label{sec:exp}

\subsection{Dataset}
Our proposed model was evaluated on the Isharah multi-scene CSLR dataset, which contains 14,000 videos from 18 signers and was processed to get the estimated 2D poses. The dataset includes 1000 unique sentences, and each pose consists of a body skeleton, hand keypoints, face keypoints, and face keypoints, totaling $86$ keypoints per frame. Specifically, the keypoint distribution consists of:
\begin{itemize}
    \item \textbf{Hand keypoints:} 42 keypoints (21 for right hand + 21 for left hand)
    \item \textbf{Face keypoint:} 19 keypoints (redundant keypoints removed)
    \item \textbf{Body skeleton:} 25 keypoints (upper body pose landmarks)
\end{itemize}

\subsection{Training details}
\textbf{Training Setup} All experiments are conducted on a single NVIDIA RTX 4090 GPU with 24GB memory. We train the model for 100 epochs with early stopping based on validation WER using a patience of 10 epochs. We adopted AdamW optimizer with an initial learning rate of $1 \times 10^{-4}$, a weight decay of $1 \times 10^{-3}$, and $\beta$ values of (0.9, 0.999). We employ a cosine annealing warm restarts scheduler with $T_0=10$, $T_{mult}=2$, and minimum learning rate of $1 \times 10^{-6}$. We use a batch size of 8 during training and 4 during inference. We normalize the input pose sequences to the range [-1, 1] and then pad or truncate them to the maximum sequence length. The model used cross-entropy loss during training.

\textbf{Data Augmentation} We applied several augmentation techniques during training: Gaussian jittering with $\sigma = 0.01$, random scaling in the range [0.85, 1.15], temporal masking with probability 0.15, frame dropout with probability 0.05, and time warping with a maximum shift of 1 frame. All augmentations are applied with a probability of 0.5.

\subsection{Evaluation Metrics}
To evaluate the performance of our approach, we used Word Error Rate (WER), which is used to measure the accuracy of the model in \textbf{sub}stituting, \textbf{ins}erting, and \textbf{del}eting text relative to the \textbf{reference} text, which is the ground truth. Where \textbf{lower} means better accuracy in generation. 

\[
\text{WER} = \frac{\#\text{sub} + \#\text{ins} + \#\text{del}}{\#\text{reference}}
\]

\subsection{Ablation Study}
\label{sec:ablation}

To conduct comprehensive ablation studies, we randomly split the original dev and test set into a new dev set (70\%) and test set (30\%). All ablation results mentioned in this section used this newly split data. 

\textbf{Effect of Feature Extraction Methods} Tab~\ref{tab:input_ablation} compares different approaches for processing temporal pose sequences. The 1D CNN with 2 layers achieves the best performance, significantly outperforming direct linear embedding. This result demonstrates that temporal compression through convolution effectively captures relevant motion patterns while reducing computational overhead. Increasing to 3 CNN layers shows diminishing returns, likely due to over-compression that loses fine-grained temporal details essential for sign recognition.

\textbf{Effect of Input Modalities} In tab~\ref{tab:modality_ablation}, we analyze the contribution of different joints of the body to recognition performance. Surprisingly, the combination of body and hands without facial features achieves the lowest WER, outperforming the full modality configuration. This result suggests that in signer-independent scenarios, facial expressions may introduce variability that hinders generalization to unseen signers. Hands-only configuration performs reasonably well, confirming that hand movements carry the primary semantic information in sign language. The poor performance of using only the hands and face keypoints indicates that body posture provides crucial contextual information that cannot be replaced by facial features alone.

\textbf{Effect of Learning Rate Scheduler}
Table~\ref{tab:scheduler_ablation} evaluates the impact of learning rate scheduling on model performance.  The results show that integrating a scheduler significantly improves the development set, reducing WER from 9.6\% to 5.4\%.  Furthermore, the scheduler allows for faster convergence.  However, the performance difference on the test set is not much, which implies that the scheduler largely improves training efficiency and stability rather than generalization ability.


\begin{table}[ht]
\centering
\caption{Ablation on input feature extraction.}
\label{tab:input_ablation}
\begin{tabular}{@{}lccc@{}}
\toprule
\multirow{2}{*}{\textbf{Input Type}} & \multicolumn{2}{c}{\textbf{WER (\%)}} & \multirow{2}{*}{\textbf{Params (M)}} \\
\cmidrule(lr){2-3}
 & \textbf{Dev} & \textbf{Test} & \\
\midrule
Linear  & 7.3 & 11.2 & 87.1 \\
1D-CNN(2 Layers)  & 5.4 & \textbf{10.1} & 87.9 \\
1D-CNN(3 Layers)  & \textbf{4.5} & 10.5 & 90.1 \\
\bottomrule
\end{tabular}
\end{table}

\begin{table}[ht]
\centering
\caption{Ablation on input modalities.}
\label{tab:modality_ablation}
\begin{tabular}{@{}p{4.0cm}ccr@{}}
\toprule
\multirow{2}{*}{\textbf{Input Modality}} & \multicolumn{2}{c}{\textbf{WER (\%)}} \\
\cmidrule(lr){2-3}
 & \textbf{Dev} & \textbf{Test} & \\
\midrule
Hands + Face (No Body)     & 8.2 & 16.5  \\
Hands Only                        & 8.3 & 16.3 \\
Body + Hands (No Face)            & \textbf{5.4} & \textbf{10.1}  \\
Full (Body + Face + Hands)        & 13.1 & 13.7 \\
\bottomrule
\end{tabular}
\end{table}


\begin{table}[ht]
\centering
\caption{Impact of learning rate scheduler on AutoSign performance.}
\label{tab:scheduler_ablation}
\begin{tabular}{@{}lccc@{}}
\toprule
\multirow{2}{*}{\textbf{Scheduler}} & \multicolumn{2}{c}{\textbf{WER (\%)}} & \multirow{2}{*}{\textbf{Best Epoch}} \\
\cmidrule(lr){2-3}
 & \textbf{Dev} & \textbf{Test} & \\
\midrule
Without Scheduler  & 9.6 & 10.4 & 47 \\
With Scheduler     & \textbf{5.4} & \textbf{10.1} & 35 \\
\bottomrule
\end{tabular}
\end{table}

\subsection{Comparison with State-of-the-Art Methods}
\label{sec:comparism}

 In this section, we present a comprehensive comparison of our approach with state-of-the-art(SOTA) methods on the Isharah-1000 dataset. During this comparison, we used the official dev and test data to compare the results. 

\begin{table}[ht]
\centering
\caption{Comparison of performance in WER (\%) on the Isharah-1000 dataset (Signer-Independent).}
\label{tab:isharah_1000_results}
\begin{tabular}{@{}lccc@{}}
\toprule
\multirow{2}{*}{\textbf{Method}} & \multirow{2}{*}{\textbf{Input}} & \multicolumn{2}{c}{\textbf{WER (\%)}} \\
\cmidrule(lr){3-4}
 & & \textbf{Dev} & \textbf{Test} \\
\midrule
VAC \cite{Min_2021_ICCV} & RGB & 18.9 & 31.9 \\
SMKD \cite{9710140} & RGB & 18.5 & 35.1 \\
TLP \cite{hu2022temporal} & RGB & 19.0 & 32.0 \\
SEN \cite{hu2023self} & RGB & 19.1 & 36.4 \\
CorrNet \cite{hu2023continuous} & RGB & 18.8 & 31.9 \\
Swin-MSTP \cite{ALYAMI2025129015} & RGB & 17.9 & 26.6 \\
SlowFastSign \cite{ahn2024slowfast} & RGB & 19.0 & 32.1 \\
\midrule
Baseline & Pose & 20.5 & 33.2 \\
\textbf{AutoSign (Ours)} & Pose & \textbf{6.7} & \textbf{20.5} \\
\bottomrule
\end{tabular}
\end{table}

 \textbf{Comparison with Video-based Methods.} Tab~\ref{tab:isharah_1000_results} demonstrates that our pose-based approach significantly outperforms existing video-based methods. The best-performing video-based method, Swin-MSTP, achieves 26.6\% WER on the test set, while our approach achieves a significant 5.1\% improved WER accuracy over it.

 This performance gain can be attributed to the elimination of background noise and environmental variations that can interfere with video-based inputs, allowing the model to focus solely on the relevant sign language movements. The superior performance across all video-based baselines validates our hypothesis that pose representations are more suitable for sign language recognition tasks, particularly in challenging signer-independent scenarios where the model must generalize to unseen signers with varying appearance characteristics.

\textbf{Comparison with Transformer + CTC Baseline.} To provide a fair comparison with other pose-based approaches, we evaluated our approach against a transformer baseline\footnote{https://github.com/gufranSabri/Pose86K-CSLR-Isharah} that was presented during the CSLR track of the multimodal sign language recognition (MSLR) challenge 2025\cite{luqman2025signeval} using the same pose inputs but with CTC loss for sequence alignment. This baseline follows the traditional CSLR pipeline: pose sequences are processed through a transformer encoder, followed by a linear classifier and CTC loss to align variable-length pose sequences with gloss sequences. Our AutoSign approach significantly outperforms this transformer + CTC baseline as seen in \ref{tab:isharah_1000_results}, demonstrating the edge in direct autoregressive generation over traditional alignment-based approaches.

\begin{table}[t]
\centering
\caption{Qualitative analysis of AutoSign and our baseline model on Signer-Independent split of Isharah-1000. Deletion, substitution, and insertion errors are highlighted in \textcolor{red}{red}, \textcolor{green}{green}, and \textcolor{blue}{blue}, respectively.}
\label{tab:qualitative}
\begin{tabular}{@{}ll@{}}
\toprule
\multicolumn{2}{c}{\textbf{Signer-Independent}} \\
\midrule
\textbf{GT:} & \RL{هو معلم لا انا مدرسه} \\
& \textit{(HE TEACHER NO I SCHOOL)} \\
\textbf{Baseline:} & \RL{هو معلم لا انا مدرسه} \\
& \textit{(HE TEACHER NO I SCHOOL)} \\
\textbf{AutoSign:} & \RL{هو معلم لا انا مدرسه} \\
& \textit{(HE TEACHER NO I SCHOOL)} \\
\midrule
\textbf{GT:} & \RL{سوال هو} \\
& \textit{(QUESTION HE)} \\
\textbf{Baseline:} & \textcolor{blue}{\RL{استفهام}} \RL{سوال هو} \\
& \textit{(\textcolor{blue}{INQUIRY} QUESTION HE)} \\
\textbf{AutoSign:} & \RL{سوال هو} \\
& \textit{(QUESTION HE)} \\
\midrule
\textbf{GT:} & \RL{استفهام هو صديق مدرسه} \\
& \textit{(QUESTION HE FRIEND SCHOOL)} \\
\textbf{Baseline:} & \RL{استفهام هو} \textcolor{red}{\RL{مدرسه}} \\
& \textit{(QUESTION HE \textcolor{red}{SCHOOL})} \\
\textbf{AutoSign:} & \RL{استفهام هو صديق مدرسه} \\
& \textit{(QUESTION HE FRIEND SCHOOL)} \\
\midrule
\textbf{GT:} & \RL{هو صديق مدرسه} \\
& \textit{(HE FRIEND SCHOOL)} \\
\textbf{Baseline:} & \RL{هو صديق} \textcolor{green}{\RL{منزل}} \\
& \textit{(HE FRIEND \textcolor{green}{HOUSE})} \\
\textbf{AutoSign:} & \RL{هو صديق مدرسه} \\
& \textit{(HE FRIEND SCHOOL)} \\
\bottomrule
\end{tabular}
\end{table}

\subsection{Qualitative Analysis of Baseline against AutoSign}

To better understand the performance differences between our baseline and AutoSign approach, we conducted a qualitative study on the dev set using the best model from each approach. 

Tab~\ref{tab:qualitative} presents representative examples where AutoSign outperforms the CTC baseline across different error types. The results highlight several key patterns: In the first example, both models successfully translate the input, proving AutoSign's accuracy in easy scenarios. However, the baseline suffers from a variety of error types, where AutoSign generates the expected gloss. In the insertion error example, the CTC baseline incorrectly inserts "INQUIRY" before the correct sequence, while AutoSign generates the exact ground truth. For deletion errors, the baseline misses "FRIEND" from the sequence, whereas AutoSign captures all words correctly. The substitution error shows the baseline replacing "SCHOOL" with "HOUSE", while AutoSign maintains the right terminology.

These examples demonstrate AutoSign's ability to generate more accurate and complete translations compared to traditional CTC-based alignment methods, illustrating the effectiveness of direct autoregressive generation for Arabic sign language recognition.

\section{Conclusion}

We propose AutoSign, a decoder-only transformer approach for pose-to-gloss translation in Arabic sign language recognition. Unlike earlier methods that rely on HMMs or CTC for sequence alignment, our approach directly generates glosses through autoregressive generation, avoiding overfitting issues and vocabulary limitations common in traditional alignment methods.

Our method relies on pre-trained AraGPT2 to improve the semantic understanding of Arabic glosses. We use pose keypoints instead of RGB frames used in earlier research to address privacy concerns and remove environmental factors, focusing solely on signer movements. This strategy is particularly beneficial for the Isharah dataset's unconstrained recording conditions.

We validate our approach on Isharah-1000, the largest SSL dataset for signer-independent recognition, achieving state-of-the-art performance with 20.5\% WER on the test set. Future work will evaluate the larger Isharah-2000 dataset and explore model compression techniques to facilitate deployment on mobile devices, matching the original data collection setup.

{
    \small
    \bibliographystyle{ieeenat_fullname}
    \bibliography{main}
}

\end{document}